\documentclass[conference]{IEEEtran}
\IEEEoverridecommandlockouts

\usepackage{verbatim}
\usepackage{multirow}
\usepackage{caption}
\usepackage{subcaption}
\usepackage{amsmath,amssymb,amsfonts}
\usepackage[linesnumbered,ruled,vlined]{algorithm2e}
\SetKwInput{KwInput}{Input}
\SetKwInput{KwOutput}{Output}
\SetKwProg{myproc}{Procedure}{}{}

\usepackage{algpseudocode}
\usepackage{mathtools}
\usepackage{listings}

\usepackage{ifpdf}
\usepackage{lmodern}
\usepackage{lipsum}
\usepackage[T1]{fontenc}
\usepackage{textcomp}
\usepackage{mdwlist}
\usepackage{pbox}
\usepackage{comment}
\usepackage{ifthen}
\usepackage{color}
\usepackage{colortbl,xcolor}
\usepackage{upgreek}
\usepackage{upgreek}
\usepackage{todonotes}
\usepackage{breakurl}

\usepackage{tikz}
\usetikzlibrary{arrows}

\usepackage[linesnumbered,ruled,vlined]{algorithm2e}
\SetKwInput{KwInput}{Input}
\SetKwInput{KwOutput}{Output}
\SetKwProg{myproc}{Procedure}{}{}

\colorlet{punct}{red!60!black}
\definecolor{background}{HTML}{EEEEEE}
\definecolor{delim}{RGB}{20,105,176}
\colorlet{numb}{magenta!60!black}

\definecolor{Gray}{gray}{0.9}
\definecolor{LightCyan}{rgb}{0.88,1,1}

\specialcomment{notes}{\begingroup \color{blue}} { \endgroup}

\newcolumntype{!}{>{\global\let\currentrowstyle\relax}}
\newcolumntype{^}{>{\currentrowstyle}}

\newcommand{\si}{\begin{enumerate}}

\newcommand{\ei}{\end{enumerate}}

\makeatletter
\let\oldfootnote\footnote
\def\footnote{\@ifstar\footnote@star\footnote@nostar}
\def\footnote@star#1{{\let\thefootnote\relax\footnotetext{#1}}}
\def\footnote@nostar{\oldfootnote}
\makeatother

\makeatletter
\renewcommand\footnotesize{%
   \@setfontsize\footnotesize\@ixpt{5}%
   \abovedisplayskip 8\p@ \@plus2\p@ \@minus4\p@
   \abovedisplayshortskip \z@ \@plus\p@
   \belowdisplayshortskip 4\p@ \@plus2\p@ \@minus2\p@
   \def\@listi{\leftmargin\leftmargini
               \topsep 4\p@ \@plus2\p@ \@minus2\p@
               \parsep 2\p@ \@plus\p@ \@minus\p@
               \itemsep \parsep}%
   \belowdisplayskip \abovedisplayskip
}
\makeatother

\lstdefinelanguage{json}{
    basicstyle=\small\ttfamily,
    numbers=left,
    numbers=none,
    stepnumber=1,
    numbersep=8pt,
    showstringspaces=false,
    breaklines=true,
    frame=lines,
    backgroundcolor=\color{background},
    literate=
     *{0}{{{\color{numb}0}}}{1}
      {1}{{{\color{numb}1}}}{1}
      {2}{{{\color{numb}2}}}{1}
      {3}{{{\color{numb}3}}}{1}
      {4}{{{\color{numb}4}}}{1}
      {5}{{{\color{numb}5}}}{1}
      {6}{{{\color{numb}6}}}{1}
      {7}{{{\color{numb}7}}}{1}
      {8}{{{\color{numb}8}}}{1}
      {9}{{{\color{numb}9}}}{1}
      {:}{{{\color{punct}{:}}}}{1}
      {,}{{{\color{punct}{,}}}}{1}
      {\{}{{{\color{delim}{\{}}}}{1}
      {\}}{{{\color{delim}{\}}}}}{1}
      {[}{{{\color{delim}{[}}}}{1}
      {]}{{{\color{delim}{]}}}}{1},
}

\DeclareGraphicsExtensions{.eps,.png}
\graphicspath{{figure/}{figure/eps/}{figure/png/}}

\makeatletter
\def\@copyrightspace{\relax}
\makeatother

\lstdefinelanguage{json}{
    basicstyle=\small\ttfamily,
    numbers=left,
    numbers=none,
    stepnumber=1,
    numbersep=8pt,
    showstringspaces=false,
    breaklines=true,
    frame=lines,
    backgroundcolor=\color{background},
    literate=
     *{0}{{{\color{numb}0}}}{1}
      {1}{{{\color{numb}1}}}{1}
      {2}{{{\color{numb}2}}}{1}
      {3}{{{\color{numb}3}}}{1}
      {4}{{{\color{numb}4}}}{1}
      {5}{{{\color{numb}5}}}{1}
      {6}{{{\color{numb}6}}}{1}
      {7}{{{\color{numb}7}}}{1}
      {8}{{{\color{numb}8}}}{1}
      {9}{{{\color{numb}9}}}{1}
      {:}{{{\color{punct}{:}}}}{1}
      {,}{{{\color{punct}{,}}}}{1}
      {\{}{{{\color{delim}{\{}}}}{1}
      {\}}{{{\color{delim}{\}}}}}{1}
      {[}{{{\color{delim}{[}}}}{1}
      {]}{{{\color{delim}{]}}}}{1},
}

\def\IEEEbibitemsep{0.2pt}

  \renewenvironment{thebibliography}[1]{%
    \begin{oldthebibliography}{#1}%
      \setlength{\parskip}{0ex}%
      \setlength{\itemsep}{0ex}%
  }%
  {%
    \end{oldthebibliography}%
  }
  
\renewcommand{\baselinestretch}{0.95}

\title{Analyzing users' sentiment towards popular consumer industries and brands on Twitter}

\author{\IEEEauthorblockN{
Guoning Hu,
Preeti Bhargava, 
Saul Fuhrmann,
Sarah Ellinger and
Nemanja Spasojevic
}
\IEEEauthorblockA{Lithium Technologies | Klout, 
San Francisco, CA\\
Email: \{guoning.hu,
preeti.bhargava,
saul.fuhrmann,
sarah.ellinger,
nemanja.spasojevic\}@lithium.com
}
}

\date{}

\begin{document}

\maketitle

\begin{abstract}
%Social media has become one of the major means of communication and content production. 
Social media serves as a unified platform for users to express their thoughts on subjects ranging from their daily lives to their opinion on consumer brands and products. These users wield an enormous influence in shaping the opinions of other consumers and influence brand perception, brand loyalty and brand advocacy. 
%Users express their thoughts freely on social media and wield an enormous influence in shaping the opinions of other consumers. These consumer voices can influence brand perception, brand loyalty and brand advocacy. 
%As a result, it is imperative that large enterprises pay more attention to mining user opinion related to their brands and products in social media communication. With social media monitoring, they will be able to tap into consumer insights to improve their quality of product, provide better service, drive sales and even identify new business opportunities. 
In this paper, we analyze the opinion of $19M$ Twitter users towards 62 popular industries, encompassing 12,898 enterprise and consumer brands, as well as associated subject matter topics, via sentiment analysis of $330M$ tweets over a period spanning a month. %In all, we analyze about 330M user-brand interactions over a period spanning a month %To the best of our knowledge, no other work has attempted to analyze sentiment of users towards different global industries and consumer brands at such a large scale. 
We find that users tend to be most positive towards manufacturing and most negative towards service industries. In addition, they tend to be more positive or negative when interacting with brands than generally on Twitter. We also find that sentiment towards brands within an industry varies greatly and we demonstrate this using two industries as use cases. In addition, we discover that  there is no strong correlation between topic sentiments of different industries, demonstrating  that topic sentiments are highly dependent on the context of the industry that they are mentioned in. We demonstrate the value of such an analysis in order to assess the impact of brands on social media. We hope that this initial study will prove valuable for both researchers and companies in understanding users' perception of industries, brands and associated topics and encourage more research in this field.
\end{abstract}

% A category with the (minimum) three required fields
%\category{J.4}{Computer Applications}{Social and Behavioral Sciences}
%\category{H.1.2}{Information Systems}{Models and Principles}[User/Machine Systems]
%\category{J.4}{Computer Applications}{Information Systems Applications}
%
%\keywords{user modeling; personalization; behavior analysis; recommended systems; online social networks; posting times;} %  Required by KDD

\section{Introduction}
\label{section:introduction}
Social media has become one of the major means of communication and content production. %An unprecedented amount of content is generated on it everyday. 
It serves as a unified platform for users to express their thoughts on subjects ranging from their daily lives to their opinion on companies and products. 
This, in turn, has made it a valuable resource for mining user opinion for tasks ranging from predicting the performance of movies to results of stock markets and elections. 
%TBD: Add citations here for all these things

Although most people are hesitant to answer surveys about products or services, they express their thoughts freely on social media and wield an enormous influence in shaping the opinions of other consumers. These consumer voices can influence brand perception, brand loyalty and brand advocacy. As a result, it is imperative that large enterprises pay more attention to mining user opinion related to their brands and products from social media. With social media monitoring, they will be able to tap into consumer insights to improve their product quality, provide better service, drive sales and even identify new business opportunities. %\footnote{\url{https://ww2.frost.com/files/8714/2366/6305/Social_Media_Customer_Engagement.pdf} accessed May 2017}. 
In addition, they can reduce customer support costs by responding to their customers through these social media channels, as 50\% of users prefer reaching service providers on social media rather than a call center \cite{ovum2014white}. 

Sentiment analysis has become synonymous with opinion mining and involves the computational study of text in order to identify sentiment polarity (positive, negative or neutral), intensity and subjectivity \cite{pang2008opinion}. It is an excellent tool for enterprises to analyze users' expressed opinions on social media without explicitly asking any questions as this approach often reflects their true opinions. Though it has drawbacks regarding the population sampled (as it may not represent the general public), it can be used to approximate public opinion.

\begin{table*}[th]
\centering
\begin{tabular}{|c|c|}
\hline
  \textbf{User brand interaction} & \textbf{Sentiment} \\
    \hline
Things I have learned today: we own way too many dishes, & \multirow{2}{*} {Negative} \\
\textbf{@comcast} is literally the WORST, and moving is the main cause of drinking problems & \\
  \hline
Beautiful morning flying out of the Bay with \textbf{@southwestair}. & \multirow{2}{*} {Positive}\\
\#baytola \#sfbay \#family \#fun \#flying https://t.co/bnogVW2X8H &    \\
  \hline
 \textbf{@united} A measure of the harm inflicted by disrupting people's travel plans:  & \multirow{2}{*} {Negative}\\
 no one here felt \$800 came close to making them whole. & \\
 \hline
 SUPER excited and grateful to be at the \#smashbox event & \multirow{2}{*} {Positive}\\
 @menloparkmall with \textbf{@sephora}! Thanks for ... https://t.co/CMexZF3pLr	& \\
  \hline
\end{tabular}
%\vspace{-0.1in}
\caption{Examples of users' interaction with a brand (highlighted) and the interaction's sentiment}
\label{table:userbrandinteractions}
\vspace{-0.1in}
\end{table*}

In this paper, we analyze the opinion of Twitter users towards different industries, encompassing several global consumer brands, and associated subject matter topics via sentiment analysis of their tweets. 
Because twitter messages are short, it is quite challenging to identify the target that an opinion is expressed towards.
As a preliminary study, here we take a special approach based on the observation that
\textbf{when a user interacts with an industry or brand, such as a reply or a retweet, the sentiment in the related text likely shows this user's opinion toward that industry or brand.}
Therefore, by aggregating such interactions from a large number of users, the sentiment distribution can reveal the general attitude about the industry or brand among twitter users.

Our dataset consists of $19M$ unique users, $62$ popular industries (such as \emph{Airlines}, \emph{Automotive}, \emph{Telecommunications} etc.), encompassing $12,898$ enterprise and consumer brands\footnote{\emph{Brand} is here defined as an independently recognizable product line or business unit.} %, often with its own logo, slogans, and so forth.} 
(such as \emph{United}, \emph{Volkswagen} etc.), 8030 subject matter topics and about $330M$ user-brand interactions over a period spanning a month. 
Specifically, we try to answer the following questions:
\begin{itemize}%[nolistsep,noitemsep]
\item What is the opinion of users towards different industries? What are the industries that are viewed negatively or positively?
%\item Are there any industries that are heavily polarized? 
\item Within industries, what is the users' sentiment towards different brands?
%\item What are the specific brands within an industry that are heavily polarized?
\item What are the frequent topics that users mention when they interact with brands over social media? What is their sentiment towards those topics?
\item How does users' general behavior on social media differ from their behavior when they interact with brands?
\end{itemize}

%Hence, large corporations have adopted social media for their marketing strategy to commercialize their products or services but also engage with their customers.  %

%Because of the democratic nature of public interactions on social networks, they provide an ideal platform to interact with brands, companies and service providers.On the other side of these interactions, many companies are developing strategies to respond to their customers on social networks.

%Consumers tend to trust the opinion of other consumers, especially those with prior experience of a product or service, rather than company marketing. Social Media are influencing consumers? preferences by shaping their attitudes and behaviors. The influence of the internet, especially via social networking, on people?s purchasing behavior has grown over the years. Monitoring the Social Media activities is a good way to measure customers? loyalty, keeping track of their sentiment towards brands or products, of the impact of campaigns and the success of marketing messages, identifying and engaging the top influencers who are most relevant to the brand, product or campaign. Social Media are the next logical marketing arena. 

Our main findings are:
\begin{itemize}%[nolistsep,noitemsep]
\item 
%Users tend to be more positive or negative when interacting with brands than generally on Twitter. In addition, users who are generally positive (negative) in their tweets tend to be positive (negative) towards brands as well, and vice versa.
There is a positive correlation between users' general sentiment and their sentiments toward brands. Users who generally express one sentiment class (positive, negative or neutral) on social media tend to express that same sentiment toward brands.
%There is a correlation between users? general sentiment and their sentiments toward brands, where a user who is generally positive tends to also be positive towards brands, and so on
\item Among industries, the most negative industries tend to be those providing services (such as \emph{Airlines}, \emph{Mail and Shipping}, and \emph{Telecommunications}) whereas the most positive industries tend to be those manufacturing and selling consumer goods such as \emph{Household appliances}. We also find that sentiment towards brands within an industry varies greatly and we demonstrate this using two industries as use cases. 
\item Finally, we find that there is no strong correlation between topic sentiments of different industries, demonstrating  that topic sentiments are highly dependent on the context of the industry that the topics are mentioned in.
\end{itemize}

To the best of our knowledge, no other work has attempted to analyze sentiment of users towards industries, consumer brands and topics at such a large scale.

%
%\section{Knowledge Base}
%\label{section:knowledgebase}
%\input{texfiles/knowledgebase}

\section{Related Work}
\label{section:relatedwork}
Several recent papers have focused on analyzing social media sentiment for a variety of purposes such as public opinion mining, stock market prediction and prediction of election results. In this paper, we do not intend to cover all work on sentiment analysis of social media exhaustively. Instead, we focus on discussing a few relevant works.

Balasubramanyan et al. \cite{Balasubramanyan2010tweets} found high correlation between public opinion (in terms of consumer confidence and political opinion), measured from polls, with sentiment measured from tweets. Hu et al. \cite{hu2013listening} analyzed public events by automatically characterizing segments and event topics in terms of the aggregate sentiments they elicited on Twitter. 
Bollen et al. \cite{bollen2011modeling} analyzed tweets to find correlation between overall public mood and social, economic and other major events. %in the media and popular culture.
Si et al. \cite{si2014exploiting} used the co-occurrences of stock tickers in tweets to propagate sentiment of stocks in order to predict stock values. 
Mittal and Goel \cite{mittal2012stock} mined public mood from Twitter and used it along with previous days' stock market values to predict future stock market movements.
Ceron et al. \cite{ceron2014every} used sentiment analysis on tweets to understand the policy preferences of Italian and French citizens. %Mihaltz et al. \cite{mihaltz2015beyond} analyzed Hungarian public Facebook comments written in response to public posts published on the pages of Hungarian politicians and political organizations in order to detect and quantify agency and communion, optimism/pessimism and individual- ism/collectivism. 
Tumasjan et al. \cite{tumasjan2010predicting} analyzed the role of political sentiment and mentions of political parties in Tweets as indicators of election results.

In contrast to this rich landscape, very few studies have focused on investigating users' sentiment towards major worldwide brands. Adeborna and Siau \cite{adeborna2014approach} propose a sentiment topic recognition model based on correlated topics models with variational expectation-maximization algorithm using airline data from Twitter. They validate their approach on a dataset of 1046 tweets concerning three major airlines by computing their airline quality rating. Mostafa \cite{mostafa2013more} used a random sample of 3516 tweets to evaluate consumers' sentiment towards 16 global well-known brands.  

%such as Nokia, T-Mobile, IBM and DHL. 
Our work differs in the scale of the Twitter dataset size as well as the number of industries and brands we analyze. %We analyze $19M$ users' sentiment towards 62 different industries as well as 11,690 global brands within these industries via $330M$ tweets. 
In addition, we look at users' sentiment towards topics. %most frequently mentioned topics. %mentioned by  with respect to these industries and brands and the sentiment towards them. 
Finally, we derive some meaningful insights from this analysis and demonstrate the value of such an analysis in order to assess the impact of brands on social media.%analyze user's general sentiment as well their sentiment when they interact with brands.

\section{Data Set}
\label{section:data_set}
%How did we  collect our data - users, industries, brands etc.
\begin{figure*}[t]
  \centering
   \begin{minipage}[b]{\textwidth}
   \includegraphics[width=\textwidth]{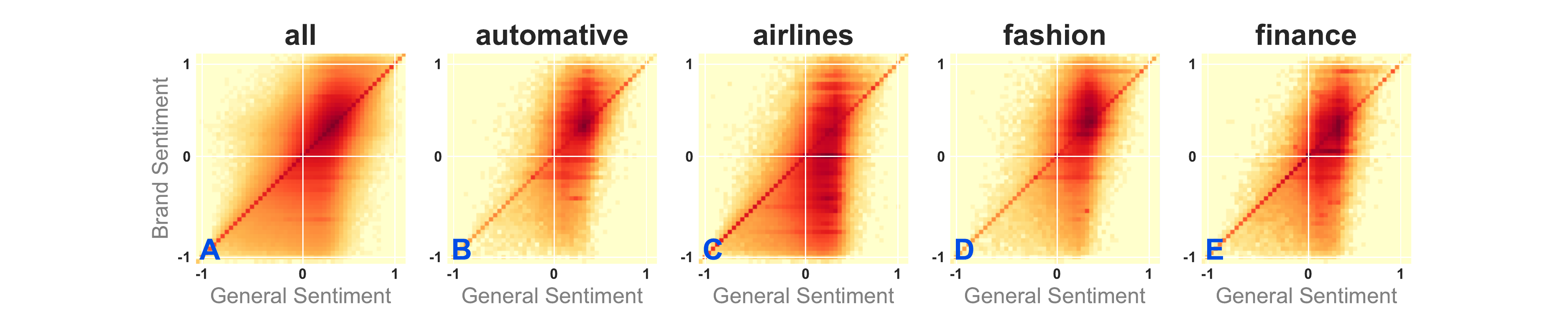}
               \vspace{-0.2in}
    \caption{Heatmap of users' brand and general sentiments}
    \label{fig:2D_histogram_sentiments}  \end{minipage}
  \begin{minipage}[b]{\textwidth}
  \includegraphics[width=\textwidth]{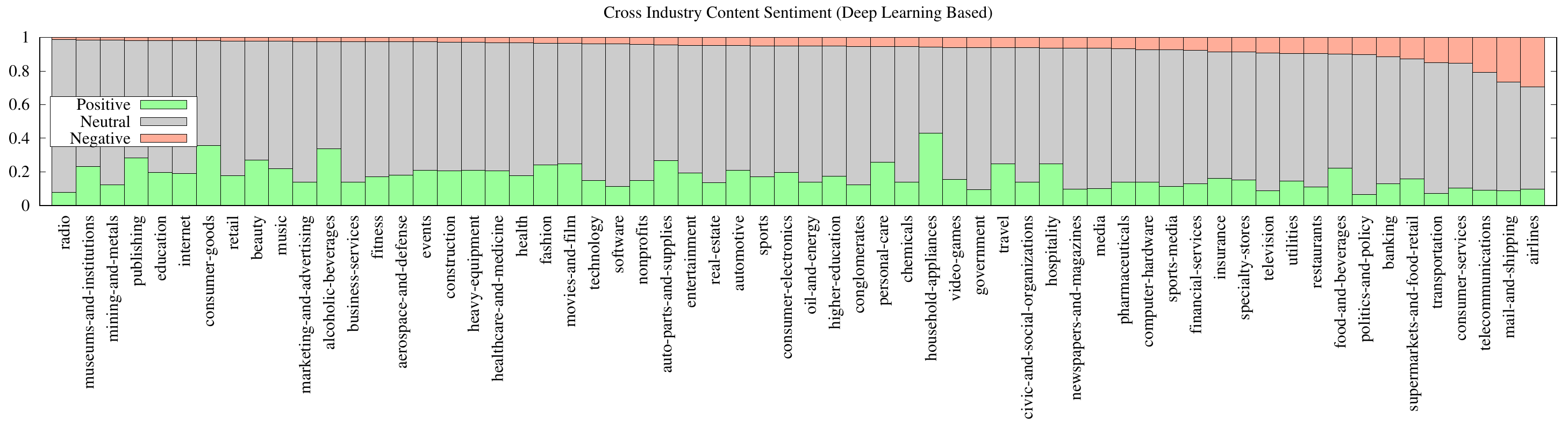}
  \vspace{-0.2in}
    \caption{Sentiment distribution by industry, ordered by negative sentiment (best viewed in color)}
    \label{fig:industry_sentiment_by_negative}
  \end{minipage}
  \begin{minipage}[b]{\textwidth}
 \includegraphics[width=\textwidth]{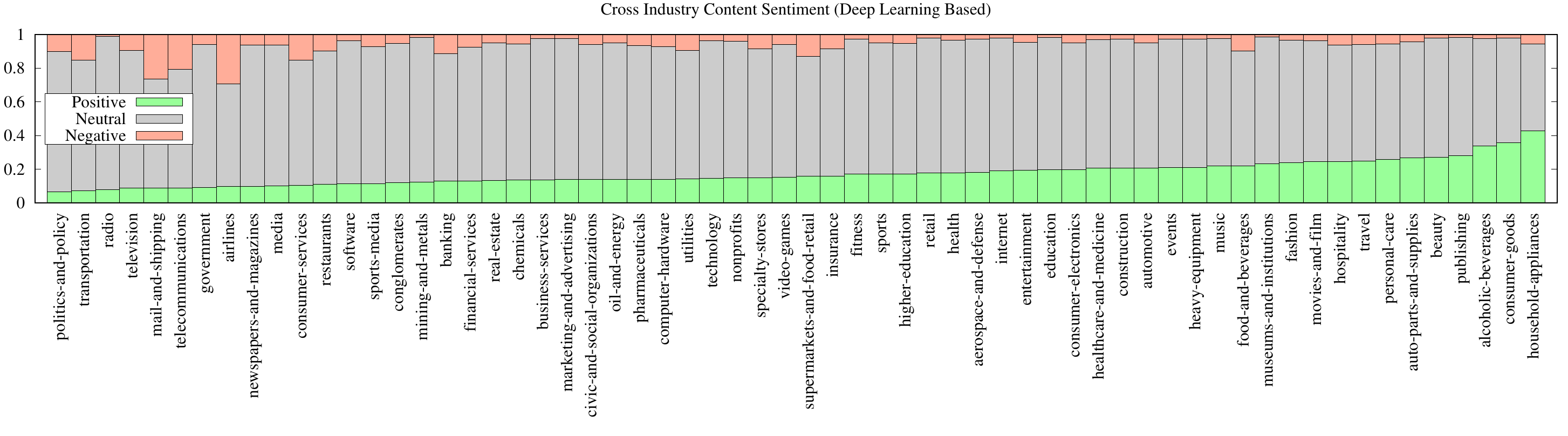}
 \vspace{-0.3in}
    \caption{Sentiment distribution by industry, ordered by positive sentiment (best viewed in color)}
           % \vspace{-0.1in}
    \label{fig:industry_sentiment_by_positive}
  \end{minipage}
  \vspace{-0.2in}
\end{figure*}

\begin{figure*}[t]
  \centering
  \begin{minipage}[b]{\textwidth}
    \includegraphics[width=\textwidth]{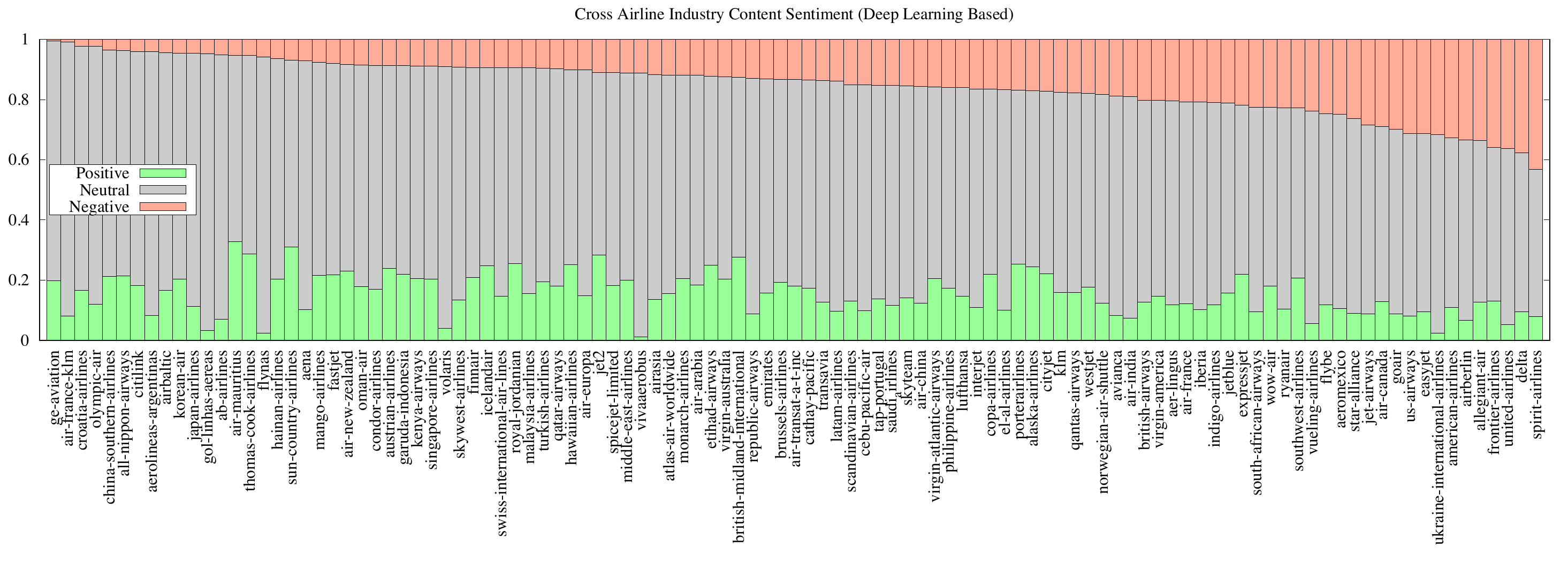}
     \vspace{-0.3in}
    \caption{Sentiment distribution for Airlines, ordered by negative sentiment (best viewed in color)}
           % \vspace{-0.1in}
    \label{fig:airline_sentiment_by_negative}
  \end{minipage}
%  \begin{minipage}[b]{\textwidth}
%    \includegraphics[width=\textwidth]{figure/png/sentiment_stacked_by_airlines_industry_by_positive_skynet.png}
%    \caption{Sentiment distribution for Airlines, ordered by positive sentiment (best viewed in color)}
%          %  \vspace{-0.1in}
%    \label{fig:airline_sentiment_by_positive}
%  \end{minipage}
% % \vspace{-0.1in}
%\end{figure*}
%
%\begin{figure*}[t]
%  \centering
%  \begin{minipage}[b]{\textwidth}
%    \includegraphics[width=\textwidth]{figure/png/sentiment_stacked_by_automotive_industry_by_negative_skynet.png}
%    \caption{Sentiment distribution for Automotive Brands, ordered by negative (best viewed in color)}
%           % \vspace{-0.1in}
%    \label{fig:automotive_sentiment_by_negative}
%  \end{minipage}
  \begin{minipage}[b]{\textwidth}
    \includegraphics[width=\textwidth]{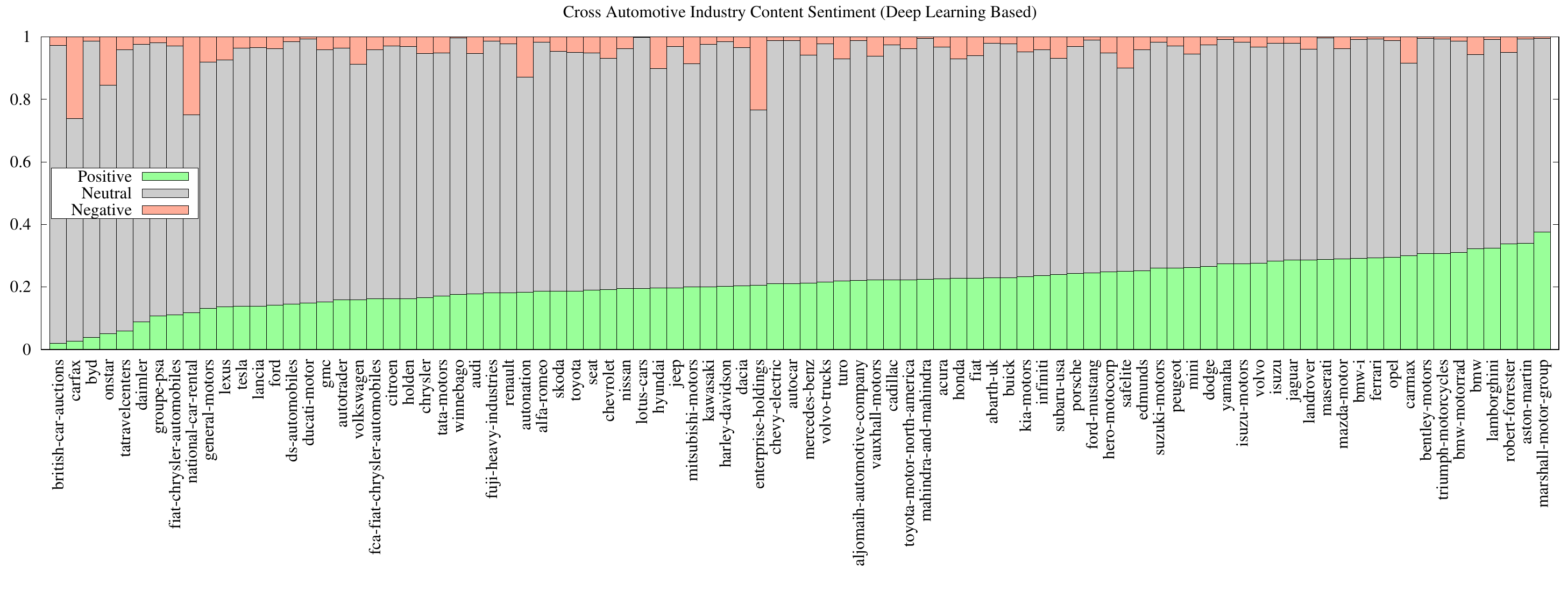}
     \vspace{-0.3in}
    \caption{Sentiment distribution for Automotive, ordered by positive sentiment (best viewed in color)}
          %  \vspace{-0.1in}
    \label{fig:automotive_sentiment_by_positive}
  \end{minipage}
  \vspace{-0.3in}
\end{figure*}

\subsection{Industries and brands dataset}

For our purposes, a brand is defined as an independently recognizable product line or business unit, often with its own logo, trademark, slogan, etc. Examples of brands in our dataset include \emph{Cadbury}, \emph{AAA}, and \emph{23andme}. A brand may be synonymous with a given company or business, but we make no attempt to represent company ownership or hierarchies (e.g. \emph{Fanta} is considered a separate brand from \emph{Coca-Cola}, although both brands are owned by the corporate entity \emph{The Coca-Cola Company}). Brands in our data set can contain one or more Twitter profiles and belong to one or more industries.

An industry is defined as a high-level category of brands that perform similar functions or provide similar products. In this analysis, we use a subset of 62 industries drawn from LinkedIn Industry Codes\footnote{http://developer.linkedin.com/docs/reference/industry-codes}, the Fortune 500\footnote{http://fortune.com/fortune500/list/}, the Forbes Global 2000\footnote{http://www.forbes.com/global2000/}, and Hoovers\footnote{http://www.hoovers.com/industry-analysis/industry-directory.html}. See Figure \ref{fig:industry_sentiment_by_negative} for a list of included industries.
 
The brand dataset was initially bootstrapped by collecting companies' names and industries from the LinkedIn profiles of Klout\footnote{https://klout.com} users. %who had opted in to LinkedIn collection. %, where the company met certain criteria. 
We then attempted to match the company names to Twitter handles via the Bing Search API. However, this approach had some limitations: %the accuracy of this approach was limited, for several reasons:
\begin{itemize}%[nolistsep, noitemsep]
\item LinkedIn included many companies that had shut down, changed names, etc.
\item Even with the industry name appended, the Bing Search API often returned non-brand Twitter accounts with similar names to the target brands.
\item Bing Search API provided only one Twitter handle per brand, whereas many brands have several specialized handles (for customer service vs. corporate communications, in different languages, etc.).
\end{itemize}
 
Thus, in order to guarantee the quality of the data, we chose to manually curate the brand dataset. Candidate handles were collected both from the initial bootstrap data and by mining the Twitter Firehose stream\footnote{http://support.gnip.com/apis/firehose/overview.html}, looking at signals such as reply volume and average time to reply (which can indicate a handle staffed by professional social media agents), follower and mention volume, and the use of enterprise-level publishing and reply tools. Each candidate handle was reviewed to determine if it belonged to a brand. If yes, it was assigned a new or existing brand node and given one or more industry labels. At this time, the brand data set contains 12,898 brands and 16,137 handles.
 
 \subsection{Twitter dataset}
 
For each brand, we collected \emph{interactions} (including replies to tweets, mentions, retweets, and cc) of Twitter users with the brand for the month of April 2017, via the Twitter Firehose stream. Each tweet was processed through the Lithium NLP pipeline \cite{Bhargava:lithiumnlp} to extract information such as the tweet language, named entities etc. For each Twitter user in our dataset, we collected all their English tweets for April 2017. 

The complete dataset contains more than $19M$ unique users that interacted with the $12,898$ brands.
These users generated more than $330M$ user-brand interactions. 
Among these interactions, there are more than $56M$ unique tweets in English.
Overall, these users posted about $282M$ unique tweets in English. Table \ref{table:userbrandinteractions} shows some examples of user brand interactions in our dataset along with their sentiment.

 \subsection{Topics dataset}\label{sec:topicsdataset}
 
Each tweet in our dataset is associated with subject matter topics that are most relevant to the tweet text. For instance, the tweet ``Anti-net neutrality bots are swarming the FCC`s comments http://engt.co/2q4Kd7l'' gets associated with topics such as  \emph{Net Neutrality}, \emph{Federal Communications Commission} and \emph{Telecommunications} etc. These topics belong to the Klout Topic Ontology\footnote{http://bit.ly/2fcg3wi} %\footnote{https://github.com/klout/opendata/tree/master/klout_topic_ontology}
(KTO) \cite{KTO} - a manually curated ontology built to capture social media users' interests and expertise scores, in different topics, across multiple social networks.  As of April 2017, it consists of roughly 8,030 topic nodes and 13,441 edges encoding hierarchical relationships among them. On Klout, we use these topics to model users' interests \cite{nemanja-lasta} and expertise \cite{Spasojevic2016:experts} by building their topical profiles. These user topics are also available via the GNIP PowerTrack API\footnote{http://support.gnip.com/enrichments/klout.html}.

The topics are inferred by mapping the recognized entities in a tweet to KTO topics via the Lithium NLP pipeline \cite{Bhargava:lithiumnlp} which uses a weighted ensemble of several semi-supervised learning models that employ entity co-occurrences, %GloVe \cite{glove2014} 
word embeddings, Freebase hierarchical relationships and Wikipedia.%\footnote{A complete description of this algorithm is beyond the scope of this paper  but we are working on publishing it.}.  %in order to propagate topic labels 

\section{Method}
\label{section:methods}
\begin{figure*}[t]
  \centering
    \includegraphics[width=\textwidth]{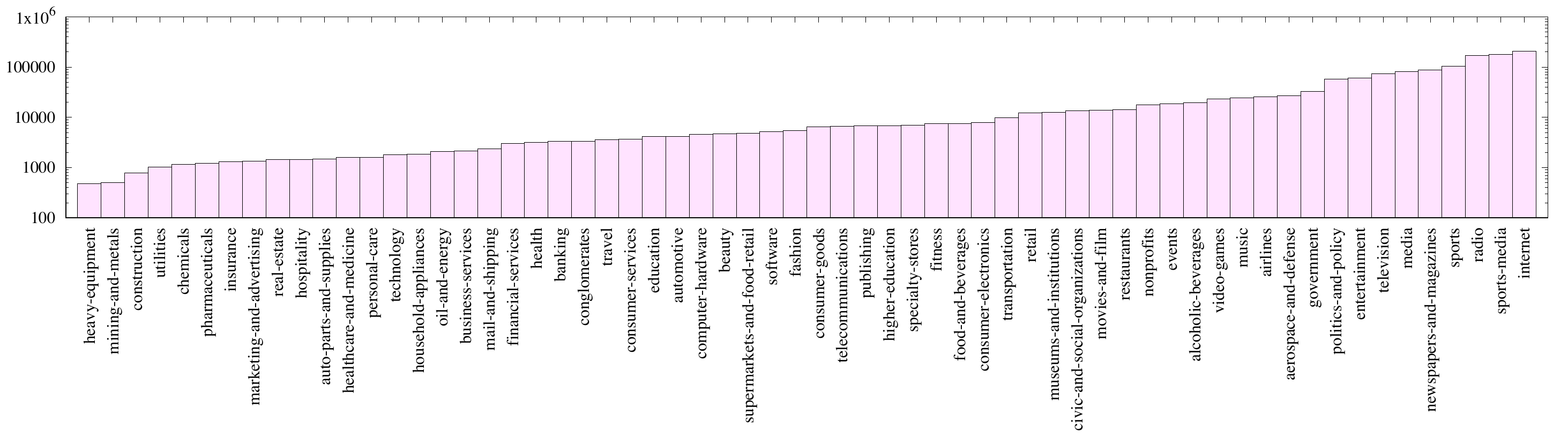}
    \vspace{-0.2in}
    \caption{Average User-Brand Interactions in Each Industry}
    \vspace{-0.2in}
    \label{fig:average_user_brand_interactions}
\end{figure*}

\begin{figure}[t]
  \centering
    \includegraphics[width=0.47\textwidth]{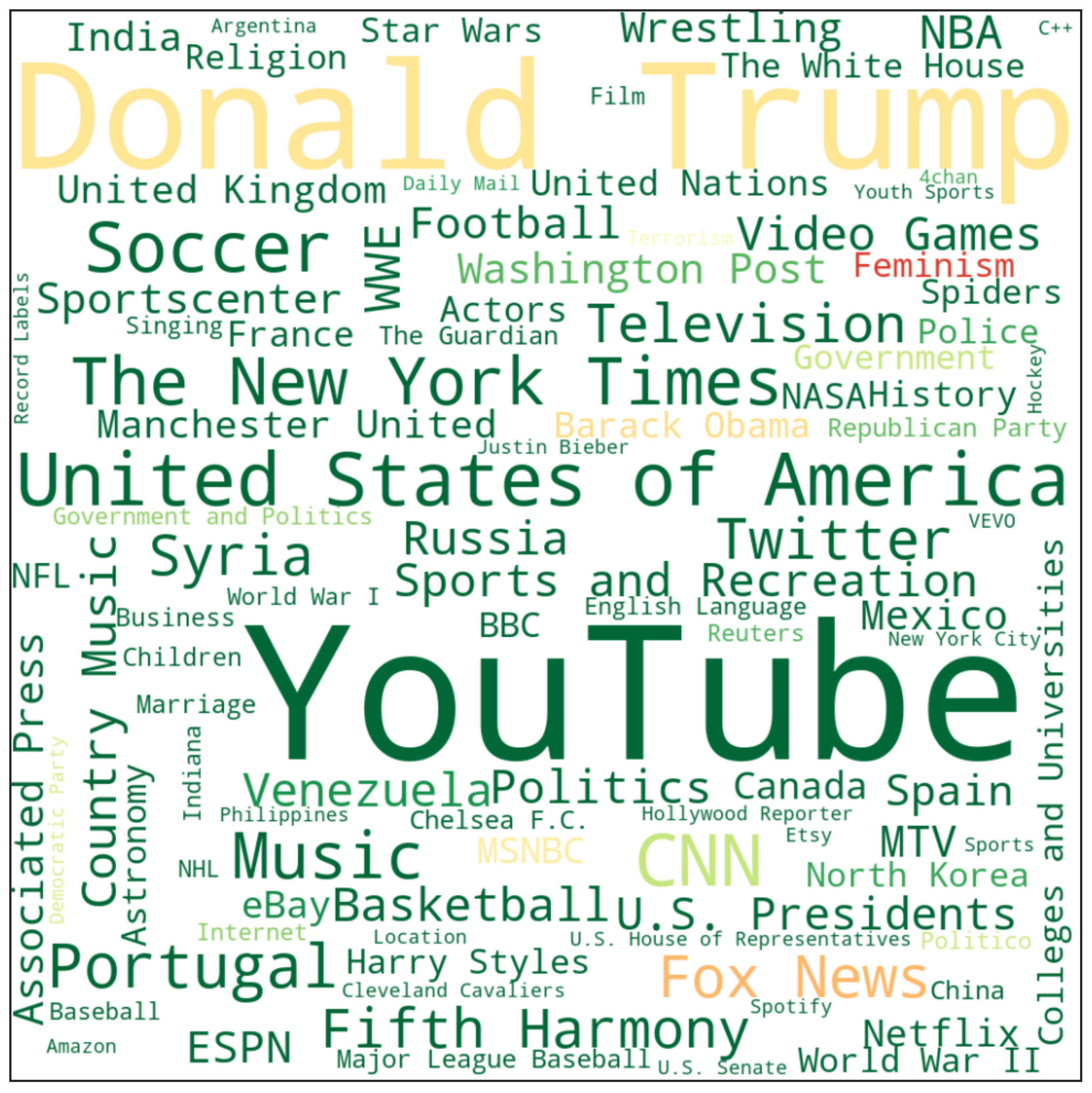}
    \vspace{-0.1in}
    \caption{Word cloud of topic sentiments across all industries with color representing the sentiment (red is negative, green is positive) and the shade representing its intensity}
            \vspace{-0.1in}
    \label{fig:word_cloud_topics}
\end{figure}

To analyze the sentiment of a tweet and label it as one of the 3 sentiment classes (`Positive', `Negative' or `Neutral'), we used a Deep Learning based approach as this field has recently enabled strong advances in NLP \cite{young:dlNLP}. 
In particular, we use Recurrent Neural Networks (RNNs) which are a class of neural networks that are able to model relationships in time. Unlike traditional neural networks, RNNs use units with internal states that can persist information about previous events. 
This makes them suitable for problems that require sequential information, such as text processing tasks. We also experimented with a lexicon based approach but found deep learning to be more accurate given empirical evidence (\cite{rao2016actionable, Spasojevic:actionability}).

In our implementation, we use a conventional Long Short Term Memory (LSTM) network \cite{rao2016actionable}. %.a form of RNN particularly optimized to learn long-term dependencies 
We first preprocess the text by tokenizing and normalizing it. We then feed this text to the LSTM, and use the labels to perform supervised learning on the messages.
We use a vocabulary of $40k$ words, $128$ embeddings units, $32$ units,
and a dropout of $0.5$. Our training dataset consists of about $14M$ message
sentiment labels that come from a brand agent manual assignment dataset \cite{Spasojevic:actionability}.
The trained model outputs a probability distribution over the 3 sentiment classes for a test sample.
To compute a final sentiment score, we use a simple approximation of taking the difference of the positive
and negative sentiment probabilities.

\section{Results}
\label{section:insights}
\begin{figure*}[t]
  \centering
    \includegraphics[width=\textwidth]{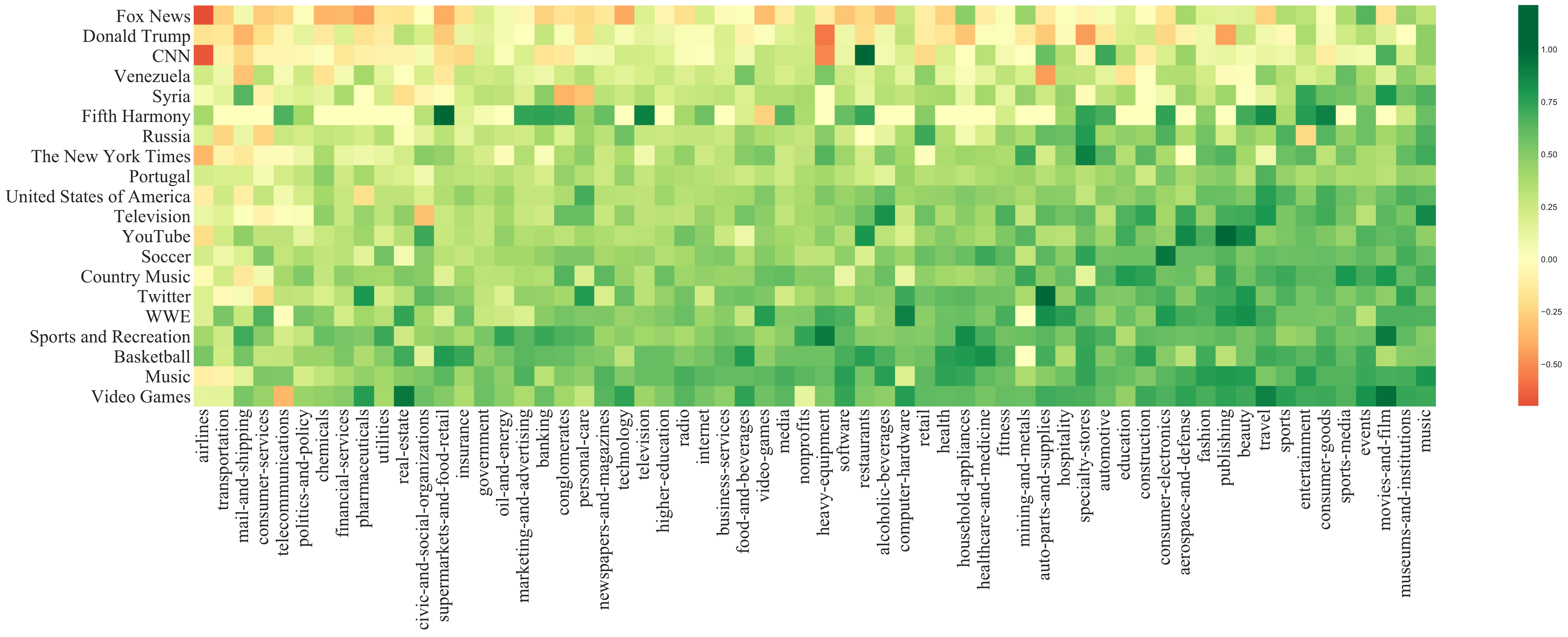}
    \vspace{-0.2in}
    \caption{Topic Sentiment across different industries (best viewed in color)}
    \vspace{-0.1in}
    \label{fig:topic_industry_co_sentiment}
\end{figure*}

\begin{figure*}[t]
  \centering
    \includegraphics[width=0.75\textwidth]{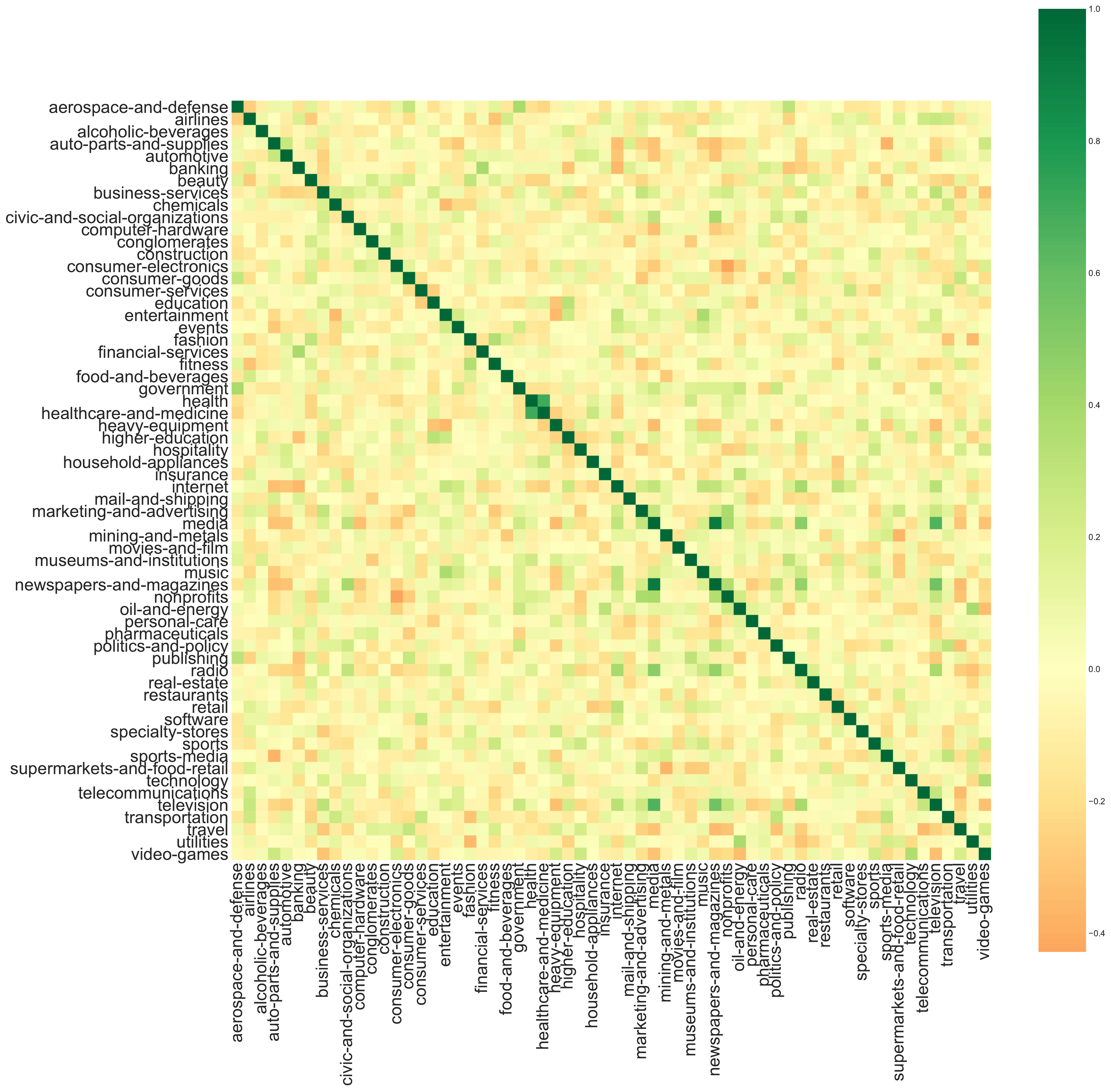}
    \vspace{-0.1in}
    \caption{Relationship between Industries represented by topic sentiment vectors (best viewed in color)}
            \vspace{-0.1in}
    \label{fig:industries_correlations}
\end{figure*}

We categorized the results obtained from this analysis into three main categories: 

\subsection{Brand sentiment vs general sentiment}
We first examined whether users showed different sentiments when interacting with brands and otherwise on Twitter.
In particular, we calculated two sentiment scores (on a scale of [-1,+1]) for each user in our dataset:

\begin{itemize}%[nolistsep, noitemsep]

\item Brand sentiment: This is the average sentiment of unique tweets posted by this user during his/her interactions with brands.
Here, we calculated the sentiment score for all brands, as well as scores for individual industries. Average sentiment is computed by taking the average of the difference of positive and negative sentiment.
%probability or sentiment?
\item General sentiment: Average sentiment of all unique tweets posted by this user (including interactions with a brand).

\end{itemize}

Figure \ref{fig:2D_histogram_sentiments}A shows the two-dimensional heatmap of these two scores for all the $19M$ users of our dataset in the 62 industries, where the brighter color indicates higher density of users. We also break down this heatmap for different industries. Figures \ref{fig:2D_histogram_sentiments}B, \ref{fig:2D_histogram_sentiments}C, \ref{fig:2D_histogram_sentiments}D, and \ref{fig:2D_histogram_sentiments}E show the histograms for \emph{Automotive}, \emph{Airlines}, \emph{Fashion} and \emph{Finance} industries respectively. 
Some important observations illustrated in these figures are:
\begin{itemize}%[nolistsep, noitemsep]
\item Both the brand and general sentiment scores are centered around 0.2.
\item General sentiment scores are more closely centered around 0.2, whereas brand sentiment scores spreads more towards 1 (positive) and -1 (negative). 
This shows that users tend to be more positive or negative than usual when interacting with brands.
\item There is positive correlation (as computed by Pearson's correlation coefficient), 0.7, between these two scores for all brands.
%In particular, the correlation between the two scores.
%This shows that users who are generally positive (negative) in their tweets tend to be positive (negative) towards brands as well, and vice versa.
This shows that users who generally express one sentiment class (positive, negative or neutral) in their tweets tend to express that same sentiment toward brands.
\item The diagonal in each figure corresponds to users whose brand sentiment score and general sentiment score are almost the same. This represents a significant proportion of all users (about 3.5\%) who do not post much other than when interacting with brands.
\end{itemize}

\subsection{Industry and brand sentiment}
We then examined sentiment distributions across different industries and brands. 
To do so, we first classified each tweet as positive, negative, or neutral based on the highest sentiment probability computed by our LSTM model. 
We then calculated the percentage of positive, negative, and neutral tweets among all unique tweets for each industry and brand.
Figure \ref{fig:industry_sentiment_by_negative} shows the percent sentiment distribution of individual industries, sorted by negative sentiment. 
%In this figure, x-axis is a list of industries, sorted by their negative percents. 
%Y-axis is a stacked bar plot of the sentiment distribution, where the top, middle, and bottom portions represent negative, neutral, and positive percents, respectively.
Figure \ref{fig:industry_sentiment_by_positive} shows the same data %as in Figure \ref{fig:industry_sentiment_by_negative}, 
except that industries are sorted by positive sentiment.

Clearly, sentiment distributions vary significantly across different industries. 
Interestingly, the most negative industries tend to be those providing services to customers, such as \emph{Airlines}, \emph{Mail and Shipping}, and \emph{Telecommunications}, whereas the most positive industries tend to be those manufacturing and selling consumer goods, such as \emph{Household appliances}. Additionally, the most polarized industries are surprisingly not \emph{Politics} and \emph{Sports} (as one might expect).

We also analyzed the sentiment within specific industries\footnote{Due to lack of space, it is not possible to show sentiment towards brands in all the 62 industries}, taking \emph{Airlines} (which has the highest negative sentiment) and \emph{Automotive} (which has a high positive sentiment) as use cases.
Figure \ref{fig:airline_sentiment_by_negative} shows the sentiment distribution of individual brands in the airline industry, sorted by negative sentiment. Clearly, sentiment distribution varies significantly across different brands. As shown, users are quite negative towards \emph{Spirit}, \emph{Delta}, and \emph{United}, while quite positive towards \emph{Air-Mauritius}, \emph{Sun-Country}, and \emph{Thomas Cook}.
%Add some more insights

Figure \ref{fig:automotive_sentiment_by_positive} show the sentiment distribution of individual brands in the \emph{Automotive} industry. Again, we observe significant variation across different brands. The most positively viewed brands include luxury brands such as \emph{Aston Martin}, \emph{Lamborghini} and \emph{BMW}, while the most negatively viewed brands include rental car companies (such as \emph{Enterprise} and \emph{National}), \emph{Hyundai} and \emph{Volkswagen}.

In addition, we noticed that brands with a higher volume of interactions tend to have more neutral sentiment.
As an illustration, Figure \ref{fig:average_user_brand_interactions} shows the average user-brand interactions across industries.
Among the top 10 industries with the most interactions, 4 of them are among the top 10 industries with the most neutral sentiment, and 8 of them are among either the bottom 10 industries with the most positive or negative sentiment.
We suspect this is because these industries get a lot of interactions and coverage from media, whose messages generally tend to be neutral.

\begin{figure*}[t]
  \centering
  \begin{minipage}[b]{0.47\textwidth}
    \includegraphics[width=\textwidth]{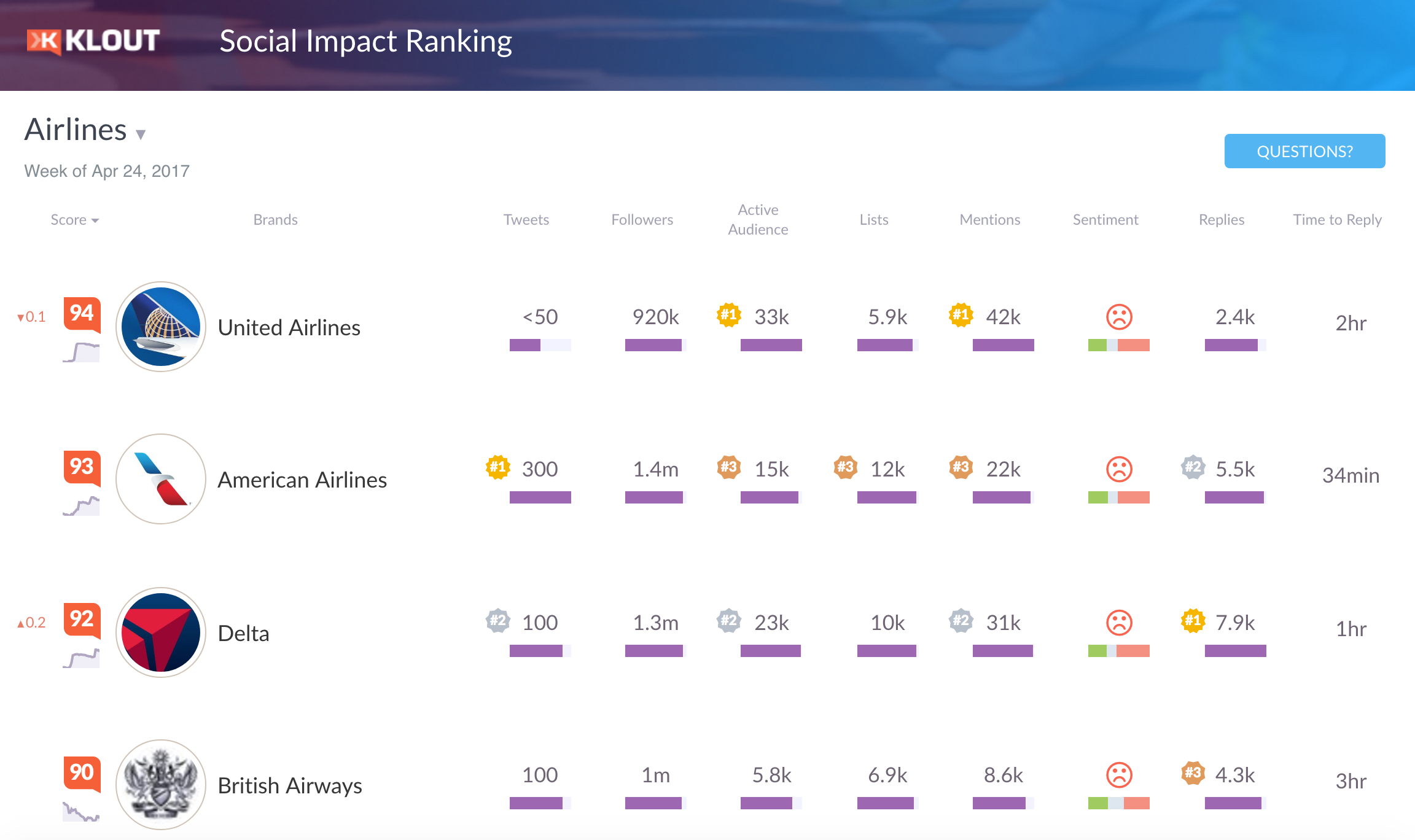}
    \caption{Airlines ranking retrieved on April 24, 2017}
            \vspace{-0.1in}
    \label{fig:airlinesold}
  \end{minipage}
  \qquad
  \begin{minipage}[b]{0.47\textwidth}
    \includegraphics[width=\textwidth]{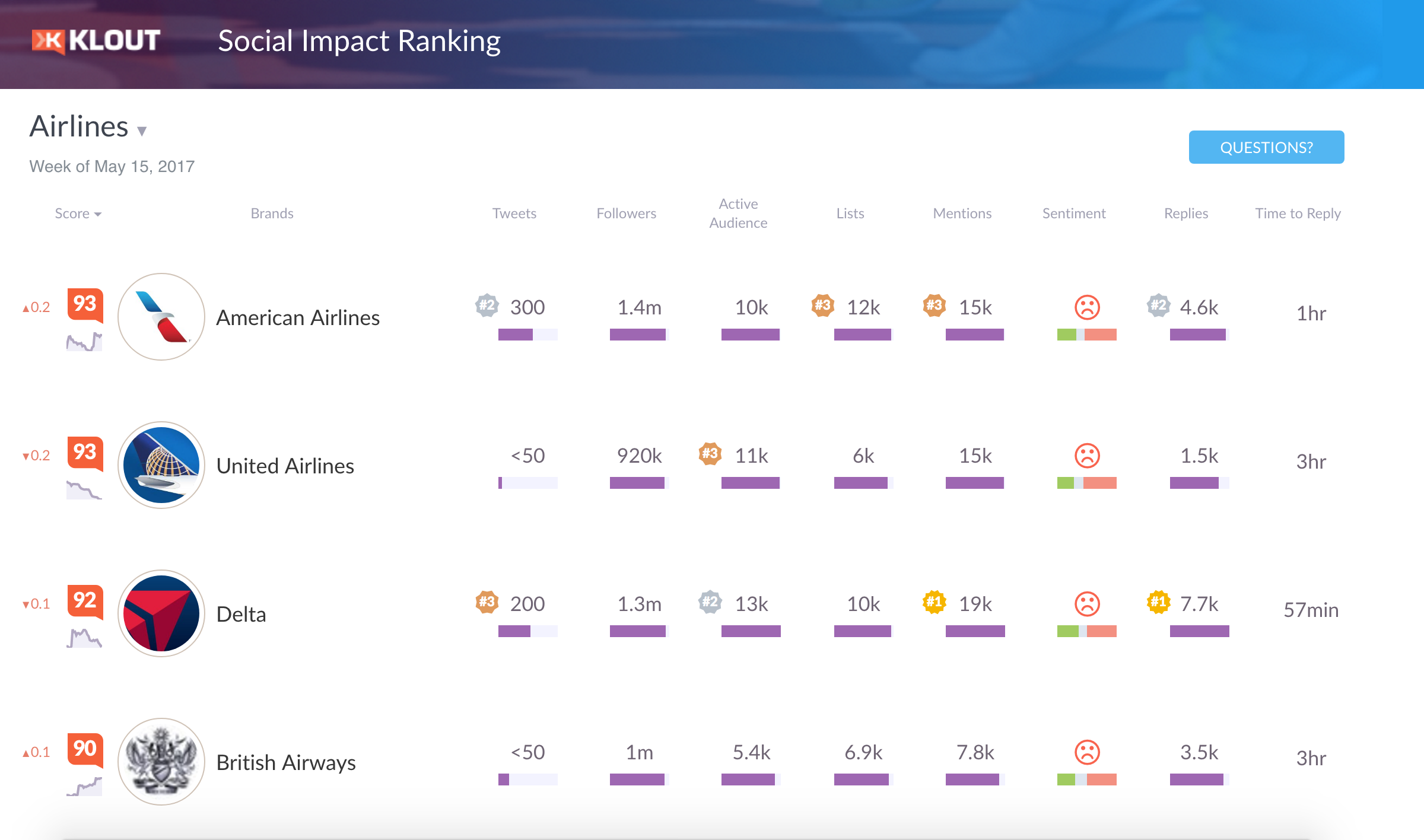}
    \caption{Airlines ranking retrieved on May 15, 2017}
                \vspace{-0.1in}
    \label{fig:airlinesnew}
  \end{minipage}
 % \vspace{-0.1in}
\end{figure*}

\subsection{Topic sentiment}
We also analyzed the sentiment towards topics %(from Section \ref{sec:topicsdataset}) 
associated with users' brand interaction tweets. Here we assume that when users interact with brands belonging to an industry, they will mention certain topics in the tweet. 
By assigning the tweet's sentiment score %(positive probability - negative probability) 
to these topics, we can roughly estimate the average sentiment of those topic within those industries.
Figure \ref{fig:word_cloud_topics} shows the word cloud  %The size of a topic is proportional to its frequency.} 
representing the average sentiment of the 100 most frequent topics in users' interactions across all industries. 
As shown, highly negative topics include \emph{Fox News} while highly positive topics include \emph{Video Games} and \emph{Music}.

Figure \ref{fig:topic_industry_co_sentiment} shows the average sentiment of the 20 most frequent topics (in users' interactions across all industries) within each industry, sorted by the average sentiment.
%In this figure, both industries and topics are .
It depicts a weak correlation between industry and topic sentiment. %he more positive or negative the overall sentiment towards it is, the more positive or negative the topic sentiment will be. 
But there are many exceptions to this, such as topic \emph{CNN} in industry \emph{Heavy Equipment} and topic \emph{Russia} in industry \emph{Entertainment}.
Moreover, average sentiments towards these topics vary significantly across different industries and are highly dependent on the context of the industry that the topics are mentioned in.
%Moreover, the topic sentiment varies within each industry. 
For instance, from Figure \ref{fig:word_cloud_topics} we can infer that \emph{Fox News} is a highly negative topic and we might expect it to be negative across all industries. However, we can see in Figure \ref{fig:topic_industry_co_sentiment} that this topic actually has positive sentiment in industries such as \emph{Events}, \emph{Music} and \emph{Sports}.

Figure \ref{fig:industries_correlations} shows the correlation between industries represented by their topic sentiment vectors. Each industry is represented as a 100 dimensional vector of the most frequent topics (those shown in Figure \ref{fig:word_cloud_topics}) and the value of each dimension represents the average sentiment towards that topic within that industry.
%where the $i$th index of the industry vector is equal to the sentiment towards the $i$th most popular topic. 
%These industry-vectors are surprisingly dense and on average contain $72$ non-zero values. 
We centered these sentiment values around zero in order to mitigate any bias in the data. %positive bias from effecting the results by creating a false sense of correlation. 
We then computed cosine similarity between the vectors.
%The simmilarity is then computed as the inner product between two zero-centered industry-vectors. 
Overall, there is no strong correlation between topic sentiments of different industries, demonstrating that topic sentiments are highly industry-dependent.

%As evident, one of the most negative topics is Strollers probably due to an incident involving strollers in American Airlines\footnote{\url{http://www.cnn.com/2017/04/22/us/american-airlines-video-confrontation-trnd/index.html}}. In addition, Air Travel and the Federal Aviation Administration (FAA) are viewed negatively. On the other hand, topics that are viewed positively include Boeing, Aviation, Babies and India. 

%\begin{figure*}[t]
%  \centering
%    \includegraphics[width=\textwidth]{figure/eps/topics_in_space2.eps}
%    \caption{Topics In Normalized Space}
%            \vspace{-0.1in}
%    \label{fig:topics_in_space2}
%\end{figure*}
%
%\begin{figure*}[t]
%  \centering
%    \includegraphics[width=\textwidth]{figure/eps/topics_in_space1.eps}
%    \caption{Topics In Normalized Space}
%            \vspace{-0.1in}
%    \label{fig:topics_in_space2}
%\end{figure*}

\section{Limitations}
\label{section:limitations}
While we have attempted to cover as much ground as possible with respect to %the number of 
industries, brands and topics as well as users' interactions and sentiment towards them, %as well as derive as many insights as possible,
we note that this study has some limitations:
\begin{itemize}%[noitemsep,nolistsep]
\item We limit our dataset to Twitter alone since it is one of the most widely used platforms for self expression. %However, we believe that similar trends will be observed when analyzing data from other social networks.
\item Our work mainly focuses on the most popular industry and brands.
%to the best of our knowledge, no other work has focussed on analyzing sentiment towards such a large scale of global industries and brands.
\item Due to resource constraints, we analyzed Twitter data for only a month.
\item The popular topics within industries are often based on current context and can change with time. % and events. %Hence, these topics
\item This study analyzed sentiment only at the message level. Accurate sentiment analysis at the aspect level will lead to more insightful results.
\end{itemize}

\section{Applications}
\label{section:applications}
As mentioned earlier, these insights about users' sentiment towards consumer brands can help brands provide better online customer service, assess the impact of social media marketing campaigns or public relations incidents, identify perceived flaws in their products, and even identify new sales or business opportunities. They also help brands gauge the consumers' perception of them. 

Another direct application of analyzing consumers' sentiment is to rank a brand relative to its competitors, allowing that brand to contextualize its performance on social media. %We have developed 
This is easily demonstrated by the Social Impact Ranking tool\footnote{https://klout.com/impact-ranking/} that Lithium has developed and which employs the sentiment of users' tweets, in addition to several other metrics, to assess the impact of different competing brands in various industries on social media. As an illustration, Figures \ref{fig:airlinesold} and \ref{fig:airlinesnew} show screenshots of the impact ranking of airlines on different dates separated by 3 weeks. Figure \ref{fig:airlinesold} coincides with an unfortunate incident involving a passenger on United Airlines\footnote{http://bit.ly/2wDUR6u}%\footnote{https://en.wikipedia.org/wiki/United_Express_Flight_3411_incident}. 
As evident, United has the highest mentions as well as strong negative sentiment towards it on Twitter. However, the number of mentions have decreased 3 weeks later (Figure \ref{fig:airlinesnew}).

Brands are very interested in understanding customers' attitudes towards them.
There have been many demands on quantitative measures of overall attitude of users towards a brand as well as their underlying business components, such as product, website, support and customer service.
Sentiment analysis of brands, topics and their relationship plays a key role in building such measures.

%\section{Evaluation}
%\label{section:evaluation}
%\input{texfiles/evaluation}

\section{Conclusion and Future Work}
\label{section:conclusion}
%Social media has become one of the major means of communication and content production. It serves as a unified platform for users to express their thoughts on subjects ranging from their daily lives to their opinion on products and companies. Users express their thoughts freely on social media and wield an enormous influence in shaping the opinions of other consumers. These consumer voices can influence brand perception, brand loyalty and brand advocacy. 
%As a result, it is imperative that large enterprises pay more attention to mining user opinion related to their brands and products in social media communication. With social media monitoring, they will be able to tap into consumer insights to improve their quality of product, provide better service, drive sales and even identify new business opportunities. 
In this paper, we analyzed the opinion of $19M$ Twitter users towards 62 popular industries, encompassing 12,898 enterprise and consumer brands, as well as associated subject matter topics via sentiment analysis of $330M$ tweets over a period spanning a month. To the best of our knowledge, no other work has attempted to analyze sentiment of users towards industries, consumer brands and topics at such a large scale. We found that users tend to be most positive towards manufacturing and most negative towards service industries. In addition, they tend to be more positive or negative when interacting with brands than generally on Twitter. We also found that sentiment towards brands within an industry varies greatly. In addition, we discovered that  there is no strong correlation between topic sentiments of different industries, demonstrating  that topic sentiments are highly dependent on the context of the industry that they are mentioned in. We also demonstrated the value of such an analysis in order to assess the impact of brands on social media using several metrics.

This preliminary study acts as a first step towards understanding the users' perception of industries, brands and associated topics. We hope that our analysis and the insights that we derive from it will prove valuable for both researchers and companies in understanding user behavior and encourage more research in this field. In future, we will work on employing these datasets results for sentiment forecasting. We also plan to conduct surveys among Twitter users in order to validate the insights we have derived from our analysis.

%\section*{Acknowledgements}
%\label{section:acknowledgements}
%\input{texfiles/acknowledgements}

%\vspace{-0.2in}
% \section{Appendix}
% \label{section:appendix}
% \input{texfiles/ds_template_appendix}

% style list: https://www.sharelatex.com/learn/Bibtex_bibliography_styles
\bibliographystyle{IEEETran}
\bibliography{bibliography}

\end{document}